\documentclass{article}

    \usepackage[preprint, nonatbib]{neurips_2021}

\usepackage[utf8]{inputenc} 
\usepackage[T1]{fontenc}    
\usepackage{hyperref}       
\usepackage{url}            
\usepackage{booktabs}       
\usepackage{amsfonts}       
\usepackage{nicefrac}       
\usepackage{microtype}      
\usepackage{xcolor}         
\usepackage{bm}
\usepackage{graphicx}
\usepackage{wrapfig}
\usepackage{lipsum}
\usepackage{booktabs}
\usepackage{subfigure}
\usepackage{amssymb}
\usepackage{xurl}
\usepackage{makecell}
\newcommand*\samethanks[1][\value{footnote}]{\footnotemark[#1]}
\title{Molecule3D: A Benchmark for Predicting 3D Geometries from Molecular Graphs}

\author{%
  Zhao Xu\thanks{These authors contributed equally to this work.}\\
  Texas A\&M University\\
  College Station, TX 77843 \\
  \texttt{zhaoxu@tamu.edu} \\
  \And
  Youzhi Luo\samethanks\\
  Texas A\&M University\\
  College Station, TX 77843 \\
  \texttt{yzluo@tamu.edu} \\
  \And
  Xuan Zhang\\
  Texas A\&M University\\
  College Station, TX 77843 \\
  \texttt{xuan.zhang@tamu.edu} \\
  \And
  Xinyi Xu \\
  Texas A\&M University \\
  College Station, TX 77843 \\
  xyxu.xd@gmail.com \\ 
  \And
  Yaochen Xie \\
  Texas A\&M University\\
  College Station, TX 77843 \\
  \texttt{ethanycx@tamu.edu} \\
  \And
  Meng Liu \\
  Texas A\&M University\\
  College Station, TX 77843 \\
  \texttt{mengliu@tamu.edu} \\
  \And
  Kaleb Dickerson \\
  Texas A\&M University\\
  College Station, TX 77843 \\
  \texttt{\small{kaleb.dickerson2001@tamu.edu}} \\
  \And
  Cheng Deng \\
  Xidian University \\
  Xi'an, 710071, China\\
  \texttt{chdeng.xd@gmail.com} \\
  \And
  Maho Nakata \\
  RIKEN \\
  Wako, Saitama 351-0198, Japan \\
  \texttt{maho@riken.jp} \\
  \And
  Shuiwang Ji \\
  Texas A\&M University\\
  College Station, TX 77843 \\
  \texttt{sji@tamu.edu} \\
}

\begin{document}

\maketitle

\begin{abstract}
Graph neural networks are emerging as promising methods for modeling molecular graphs, in which nodes and edges correspond to atoms and chemical bonds, respectively. Recent studies show that when 3D molecular geometries, such as bond lengths and angles, are available, molecular property prediction tasks can be made more accurate. However, computing of 3D molecular geometries requires quantum calculations that are computationally prohibitive. For example, accurate calculation of 3D geometries of a small molecule requires hours of computing time using density functional theory (DFT). Here, we propose to predict the ground-state 3D geometries from molecular graphs using machine learning methods. To make this feasible, we develop a benchmark, known as Molecule3D, that includes a dataset with precise ground-state geometries of approximately 4 million molecules derived from DFT. We also provide a set of software tools for data processing, splitting, training, and evaluation, etc. Specifically, we propose to assess the error and validity of predicted geometries using four metrics. We implement two baseline methods that either predict the pairwise distance between atoms or atom coordinates in 3D space. Experimental results show that, compared with generating 3D geometries with RDKit, our method can achieve comparable prediction accuracy but with much smaller computational costs. Our Molecule3D is available as a module of the MoleculeX software library (https://github.com/divelab/MoleculeX).
\end{abstract}

\section{Introduction}

In recent years, significant progress has been made in data-driven molecular representation learning methods. Many methods \cite{yang2019analyzing,stokes2020deep,wang2020advanced} use low-dimensional representations of molecules as the inputs, such as SMILES \cite{weininger1988smiles} sequences and molecular graphs. On the other hand, the ground-state 3D geometries of molecules, i.e., the 3D coordinates of all atoms in the ground-state molecule, are critical to a variety of applications, including but not limited to analysis of the molecular dynamics, prediction of biological activities of molecules, and design ligands or 3D linkers \cite{imrie2020deep, yang2020syntalinker} for proteins. Existing studies \cite{wu2018moleculenet, gilmer2017neural} also demonstrate that use of ground-state 3D geometries as inputs of machine learning models can improve the prediction performance of quantum properties. However, obtaining the ground-state 3D geometry of a molecule can take hours of computation with density functional theory (DFT). A promising way to reduce the computational cost is predicting ground-state 3D geometries from input molecular graphs by using machine learning models. Nonetheless, no benchmark dataset has been specifically designed for this geometry prediction task currently. Meanwhile, due to the poor accuracy of current fast geometry generative methods, no prior works employ the generated 3D geometries for downstream applications.

In this work, we propose a novel benchmark known as Molecule3D, which is the first benchmark that enables the systematic study of the ground-state 3D molecular geometry prediction task. In Molecule3D, we curate a large dataset by collecting and filtering molecules from PubChemQC \cite{nakata2017pubchemqc}, a chemical database developed for quantum chemistry study. The dataset contains the information of over 3.9 million molecules, including their molecular graphs, ground-state 3D geometries, and various quantum properties. We develop a set of software tools to process the dataset and perform the 3D geometry prediction task. We design 4 metrics to evaluate the geometry prediction performance. In addition, to verify the benefits of the predicted 3D geometries, we employ a test bed based on the quantum property prediction task. Concretely, we use 3D graph networks to predict the provided quantum properties based on the predicted 3D geometries and compare the prediction performance with that of using solely molecular graphs.

In addition to the dataset, we also provide a few baseline methods in the Molecule3D benchmark. We develop a deep graph neural network model for the ground-state 3D geometry prediction. Specifically, our model is based on DeeperGCN \cite{li2020deepergcn} and DAGNN \cite{liu2020towards}. It predicts the pairwise distances between all atoms in the molecule. We show that our baseline method can achieve comparable prediction accuracy with the ETKDG \cite{riniker2015better} method in the RDKit \cite{rdkit} package. In addition, we use the SchNet \cite{schutt2017schnet} model to predict the HOMO-LUMO gap property of the molecule from the ground-state geometry computed by the trained baseline model. The property prediction performance of using the predicted 3D geometry is compared with that of solely using molecular graphs. We anticipate that our work on Molecule3D would open up new avenues for improving molecular simulation and analysis using 3D geometries.

\section{Related Work}

Graph neural networks \cite{kipf2017semi, velickovic2018graph,xu2018how,gao2018large, gao2019graph} (GNNs) have emerged as attractive and powerful methodologies for representation learning from data with relational structures. Recently, GNNs have been widely studied in molecular property prediction problems. A lot of GNN architectures \cite{duvenaud2015convolutional, kearnes2016molecular, gilmer2017neural, yang2019analyzing, wang2020advanced} have been proposed for representation learning and property prediction from molecular graphs. To evaluate and compare various molecular graph representation learning methods, several molecule benchmarks, such as MoleculeNet \cite{wu2018moleculenet} and TDC \cite{huang2021therapeutics}, have been developed. In addition to molecular graphs, the accurate prediction of some quantum properties, such as HOMO-LUMO gap \cite{hu2021ogb, liu2021fast, ying2021first,addanki2021large}, requires the ground-state 3D molecular geometries as inputs. Recently, several novel 3D graph neural networks \cite{schutt2017schnet, unke2019physnet, klicpera2019directional, liu2021spherical} have been proposed to incorporate the 3D positional information of atoms for learning molecular representations.

While 3D geometries are generally useful in many molecular analysis tasks, obtaining such information is highly nontrivial and requires either extensive experiments or expensive computations through density functional theory (DFT). The choice of ground-truth labels for quantum properties also depends either on DFT computation or on experimentally collected results. Note that DFT-derived labels will typically have some bias relative to experimentally determined labels but are easier to acquire than label acquisition in the laboratory. Some machine learning studies \cite{thawani2020photoswitch, hey2020machine} use experimentally determined labels, while some others rely on DFT-derived labels.
 
Recently, some studies \cite{simm2020generative, mansimov2019molecular, xu2021an, shi2021learning, xu2021learning, ganea2021geomol} propose to efficiently generate multiple 3D geometries of a molecule with deep generative models. Accordingly, the GEOM \cite{axelrod2020geom} dataset has been created as a benchmark for this 3D geometry generation task and provides 3D geometries for over 430,000 molecules simulated by the first principle. In this dataset, a single molecule is associated with multiple low-energy 3D geometries, but they are non-ground-state geometries. Generative methods consider a one-to-many problem, i.e., generating multiple non-ground-state molecular geometries of the given molecular graph. To obtain the quantum property of a molecule, generative methods compute and average the properties of all generated molecular geometries. However, many works of 3D GNN [cite] have already proved that merely using the ground-state geometry with the lowest total energy can significantly boost the performance of the property prediction task. Hence, we propose a different, one-to-one method, that is, directly predicting the unique ground-state geometries of molecules. Compared to generation approaches targeting multiple non-ground-state geometries, prediction approaches targeting the ground-state geometries are interpreting and solving the geometry problem from a distinct perspective. We believe prediction approaches can also be effective, efficient, and promising to construct 3D geometries.


Despite of their importance, there are only very limited benchmark datasets containing ground-state 3D molecular geometries currently. The Atom3D \cite{townshend2020atom3d} benchmark is a collection of multiple diverse datasets, including QM9 \cite{ramakrishnan2014quantum}, for property prediction based on 3D molecular geometries. Except for QM9, all datasets in Atom3D are collections of large compounds, such as proteins or RNA, which are too diverse to be used in predicting 3D geometries. QM9 is a widely used dataset for quantum property prediction and includes over 130,000 molecules together with their ground-state 3D geometries obtained by DFT computations. However, molecules in the QM9 dataset are very simple. The number of heavy atoms are limited to less than 9, and types of all atoms are restricted to be one of hydrogen, carbon, nitrogen, oxygen, and fluorine. The QM7/QM7b and QM8 datasets from the MoleculeNet benchmark have the same constraints as QM9, and they even have fewer data than QM9. In summary, all current 3D molecule datasets are designed for property prediction tasks, and they are not suited to be adapted to 3D geometry prediction tasks. Given the limitations of existing benchmarks, a sufficiently large and AI-ready benchmark specifically designed for ground-state 3D molecular geometry prediction is desired.

\section{The Proposed Dataset}
We describe the proposed Molecule3D dataset in this section, including motivations and significance of the 3D geometry prediction task, details on the dataset and its source, data splits, and evaluation metrics.

\subsection{Significance of 3D Geometry Prediction}
The relationships between properties or activities of a chemical compound and its structure are widely studied since a molecule's structure is recognized as critical to determining its physical, chemical, and biological behaviors. Drug discovery and materials informatics often involve quantitative structure-property relationship (QSPR) and quantitative structure–activity relationship (QSAR). Both QSPR and QSAR are under the assumption that structurally similar molecules exhibit similar properties and activities. In these studies, the structural information of molecules is represented in various forms. For example, Faber-Christensen Huang-Lilienfeld (FCHL) \cite{christensen2020fchl} and smooth overlap of atomic densities (SOAP) \cite{bartok2013representing} are two representations that describe atomic environments and they have been used by quantum machine learning community for a long time. One-dimensional (1D) structural keys is another representation to describe molecular composition and structures as a Boolean vector. Recently, it becomes popular to use two-dimensional (2D) molecular graphs representing atoms/bonds as nodes/edges to show connections explicitly. However, the specific shape of a molecule still lacks in 2D graphs. 3D molecular geometries provide additional spatial information, and recent studies show that using 3D geometries leads to more accurate molecular property prediction than using 2D graphs only. This improvement is consistent with the fact that properties are often closely related to the shape of a molecule \cite{engel2018applied}.

Despite the importance of molecular structures, 3D geometries are challenging and expensive to construct. Currently, quantum chemistry calculations such as Gaussian, QChem, MolPro, NWChem, Turbomole, and GAMESS are adopted as common methods for computing accurate 3D geometries of molecules. Compared to expensive experimental approaches, these quantum chemistry calculations are more economical. Even so, it is a time-consuming process to obtain 3D molecular geometries. For example, obtaining the precise 3D geometries of a small molecule using DFT requires hours of computations. Currently, the high computational cost seriously impedes the wide usage of accurate 3D geometries in molecular analysis and simulation.

\subsection{Overview of Molecule3D}

\begin{figure}
    \centering
    \includegraphics[width=0.83\textwidth]{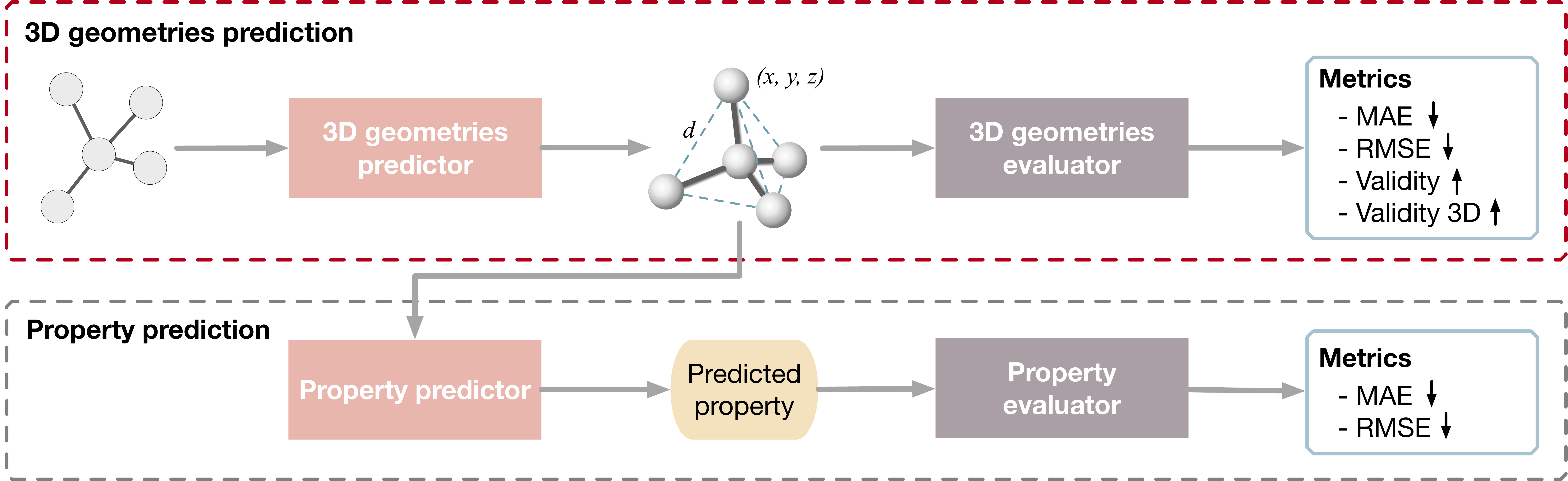}
    \caption{An overview of the Molecule3D benchmark. Given a molecular graph, we predict its ground-state geometry by employing a 3D geometries predictor and then evaluate the performance by four metrics. The predicted 3D geometry can be further used to benefit molecular property prediction.}
    \label{fig:overview}
\end{figure}

In this work, we take a fundamentally different approach to obtain 3D molecular geometries. We make use of recent advances in deep learning models that are able to capture highly nonlinear and complex relations between inputs and outputs. We propose to predict the 3D geometries of molecules using their molecular graphs as inputs. Such a machine learning approach requires the construction of training dataset consisting of matched pair of molecular graphs and corresponding 3D geometries. To this end, we propose a benchmark, known as Molecule3D, in this work. Molecule3D consists of a dataset with molecular graphs and 3D geometries, baseline methods, evaluation metrics, and sample downstream tasks using the predicted geometries. With Molecule3D, deep learning models can be trained to learn the relationship between matched pairs of molecular graphs and geometries. Once 3D molecular geometries can be predicted accurately and efficiently, they can be used directly in many downstream tasks such as property prediction. Predicting properties will benefit a wide range of real-world applications such as molecular discovery or virtual screening campaigns. For example, using predicted 3D molecular geometries as features can help predict HOMO, an essential property in discovering new electron acceptor materials in organic photovoltaics. Besides this application in renewable energy technologies, the predicted 3D geometries can also be used in drug discovery open challenge MIT AI Cures \cite{aicures}. This challenge aims to discover new antibiotics to cure the secondary lung infections caused by COVID-19, but 3D molecular geometries are not provided in AI Cures. Thus, almost all participants are using molecular graphs, SMILES sequences, or fingerprints. With the Molecule3D dataset and the provided pipeline, one can train a model to predict relatively accurate ground-state geometry.  Then, this pre-trained model with knowledge of 3D geometries can help predict antiviral activity against SARS-CoV-2. In short, we believe 3D geometry prediction has a potential to provide a new avenue for the computation of 3D geometries, thereby overcoming the current dilemma of scarce 3D geometries data. The proposed Molecule3D is anticipated to lower the barrier of using valuable 3D geometries, thereby advancing studies and improving the performance of structure-related tasks and their real-world applications.

Note that a molecule may correspond to multiple spatial arrangements of atoms known as conformations. Generally, every conformation, or equivalently molecular geometries, corresponds to a state of specific potential energy. For example, at the equilibrium ground-state, a molecule has the lowest possible energy due to the minimized repulsion and maximized attraction caused by the conformation with the most spread electron pairs. In this work, we propose to consider a molecular graph and its most stable geometry present at the ground state. 


\subsection{The Molecule3D Dataset}
The PubChemQC database is a large-scale public database containing 3,982,751 molecules with 3D geometries. First, PubChemQC extracts the IUPAC International Chemical Identifier (InChI) and Simplified Molecular Input Line Entry Specification (SMILES) from its source, the PubChem database \cite{kim2016pubchem} maintained by the National Institutes of Health (NIH). Then, PubChemQC calculates both ground-state and excited-state 3D geometries of molecules using DFT at the B3LYP/6-31G* level. To expedite the calculation, the authors of PubChemQC use RICC supercomputer (Intel Xeon 5570 2.93 GHz, 1024 nodes), QUEST supercomputer (Intel Core2 L7400 1.50 GHz, 700 nodes), HOKUSAI supercomputer (Fujitsu PRIMEHPC FX100), and Oakleaf-FX supercomputer (Fujitsu PRIMEHPC FX10, SPARC64 IX 1.848 GHz). Despite these powerful computation resources, only several thousand molecules can be processed per day. Therefore, it takes years of efforts to create such an extensive database with accurate 3D molecular geometries.

\begin{wraptable}{r}{0.65\textwidth}\vspace{-0.5cm}
    \small
    \centering
    \caption{Statistics of the Molecule3D dataset.}
    \begin{tabular}{lc}
    \toprule
    $\#$Total molecules & 3,899,647 \\
    $\#$Molecules in training set& 2,339,788 \\
    $\#$Molecules in validation set& 779,929 \\
    $\#$Molecules in test set& 779,930 \\
    Splits ratio & 6:2:2 \\
    Split type & Random/Scaffold \\
    $\#$Total atoms per molecule & 29.11 \\
    $\#$Heavy atoms per molecule & 14.08 \\
    $\#$Bonds per molecule & 29.53 \\
    Metrics & MAE, RMSE, validity, validity3D \\
    \bottomrule
    \end{tabular}
    \label{tab:stat}
\end{wraptable}

PubChemQC provides both ground-state and excited-state 3D molecular geometries. However, the raw data of PubChemQC can not be easily accessed and is not ready for machine learning. After intensive communication with authors of PubChemQC, we propose the Molecule3D that primarily serves to develop deep learning methods for predicting the ground-state 3D geometries. We firstly build a list of valid molecules by removing all molecules with invalid molecule files, with SMILES conversion error and RDKit warnings, with sanitize problem, or with damaged log files. This finally yields 3,899,647 molecules with corresponding ground-state 3D geometries. We combine all valid molecule files into four SDF files, and each file stores 1 million molecules. Moreover, we parse log files using cclib package \cite{o2008cclib} and extract four essential quantum properties for every molecule, including energies of the highest occupied molecular orbital (HOMO) and the lowest unoccupied molecular orbital (LUMO), the HOMO-LUMO gap, and total energy. Extracted properties are saved in a CSV file with the same order of the valid molecule list. To further facilitate the use of 3D geometry data in deep learning, we process the raw data and provide atom features, bond features, and SMILES strings in addition to 3D coordinates. We encapsulate all these information in a PyTorch Geometric dataset. Technical details about data processing procedure are described in Appendix~\ref{sec:data_details}. Statistics of the proposed Molecule3D are given in Table~\ref{tab:stat}.

\subsection{Data Splits}\label{subsec: splits}

\begin{wrapfigure}[12]{R}{9cm}\vspace{-0.5cm}
  \begin{center}
    \includegraphics[width=0.6\textwidth]{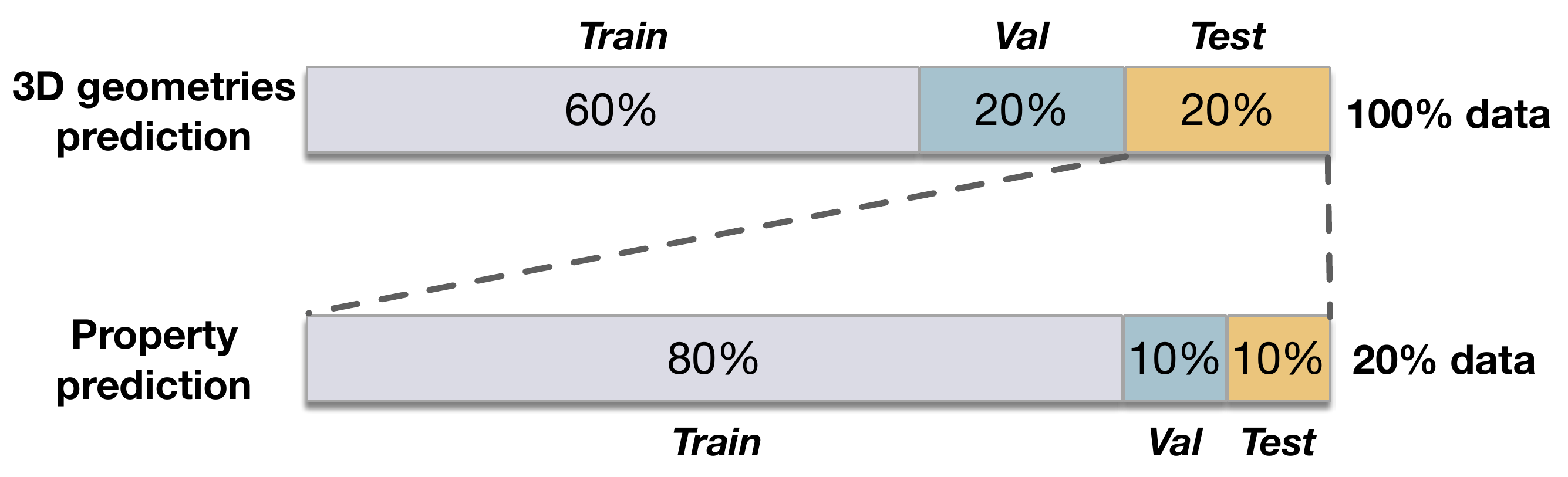}
  \end{center}
  \caption{Data splits. For both random and scaffold split, 3D geometries prediction task involves the entire dataset, while downstream property prediction task uses only 20\% of data.}
  \label{fig:splits}
\end{wrapfigure}

The entire Molecule3D dataset is split into $60\%/20\%/20\%$ for training/validation/test. We can train models for 3D geometry prediction on the training set and choose hyperparameters based on the validation set. Afterward, predicted 3D geometries on the test set can be used to evaluate the model performance. The 20\% test set is further split into smaller training, validation, and test sets with the ratio of 8:1:1, which is used for downstream property prediction tasks. The illustration of data splits is shown in Figure\ref{fig:splits}.

Together with the Molecule3D dataset, we provide two split types, random split and scaffold split. Random split ensures training, validation, and test data are sampled from the same underlying probability distribution. A molecular scaffold refers to a molecule's core component consisting of connected rings without any side chains. Splitting the dataset based on molecular scaffolds leads to a distribution shift between training and test datasets. Thus, scaffold split forces the model to capture such distribution shifts in chemical space and measures the out-of-distribution generalization ability of the model.

\subsection{Evaluation Metrics}
The 3D geometry prediction problem can be formulated as predicting the 3D coordinates of atoms or predicting the pairwise distances between atoms. While there exist infinite sets of correct 3D coordinates for a given molecule, the pairwise distances are invariant to the translation and rotation of the molecule. Hence, we propose to evaluate the performance of the 3D geometry prediction task with the error of predicted pairwise distances. When predicting 3D coordinates, we can calculate the pairwise distances from predicted 3D coordinates for evaluation.

In addition to the error of predicted pairwise distances, we propose to evaluate predicted geometries from another perspective, that is, validity. Specifically, we evaluate whether the predicted distance matrix is a valid Euclidean distance matrix (EDM), which can be used to successfully reconstruct atom coordinates in 3D space. 

\subsubsection{MAE and RMSE}
Given a dataset of totally $N$ intra-molecular pairwise distances between atoms, the ground-truth distance labels are denoted as $\{\hat{d_i}\}_{i=1}^N$ and the predicted distances are $\{d_i\}_{i=1}^N$, where $\hat{d}_i, d_i \in \mathbb{R}$. The first evaluation metric is mean absolute error (\textsc{MAE}) \cite{willmott2005advantages}:
\begin{equation}\label{eqn:mae}
    \textsc{MAE}\left(\{\hat{d}_i\}_{i=1}^N,\{d_i\}_{i=1}^N\right)=\frac{1}{N}\sum_{i=1}^N |\hat{d}_i-d_i|.
\end{equation}

The second evaluation metric is root-mean-square error (\textsc{RMSE}):
\begin{equation}\label{eqn:rmse}
    \textsc{RMSE}\left(\{\hat{d}_i\}_{i=1}^N,\{d_i\}_{i=1}^N\right)=\sqrt{\frac{1}{N}\sum_{i=1}^N \left(\hat{d}_i-d_i\right)^2}.
\end{equation}

\subsubsection{Validity of Distance Matrix and 3D Coordinates}
\label{sec:validity}

An EDM stores squared Euclidean distances between a set of points. Given a molecule of $n$ atoms, its ground-truth EDM is denoted as $\hat{D} \in \mathbb{R}^{n \times n}$. The squared distance is given as $\hat{D}_{ij}=\|\boldsymbol{\hat{p}_i}-\boldsymbol{\hat{p}_j}\|_2^2$, where $\boldsymbol{\hat{p}_1}, ..., \boldsymbol{\hat{p}_n} \in \mathbb{R}^3$ represent the ground-truth atom positions in 3D space. The predicted distance matrix is denoted as $D \in \mathbb{R}^{n \times n}$. Note that a predicted distance matrix may not be a valid EDM. In this case, we cannot obtain coordinates of atoms from the predicted distance matrix. Therefore, we further propose to use the percentage of valid EDM to evaluate predicted geometries.


According to the theorem of Gower \cite{gower1982euclidean}, the distance matrix $D$ is a valid EDM if and only if $-\frac{1}{2}JDJ$ is positive semi-definite, where $J=I-\frac{1}{n} \bm 1 \bm 1^T$ is the geometric centering matrix and $\bm 1=(1,...,1)^T \in \mathbb{R}^n$. To reconstruct atoms positions, we leverage the relationship between $D$ and its Gram matrix $M \in \mathbb{R}^{n \times n}$, that is:
\begin{equation}
    M_{ij} = \|w_i-w_j\|_2^2 = \frac{1}{2} \left(D_{1j}+D_{i1}-D_{ij}\right).
\end{equation}
$\{\boldsymbol{w}\}_{k=1}^n$ denotes a set of direction vectors from the first atom to all atoms, and $\boldsymbol{w_k}=\boldsymbol{p_k}-\boldsymbol{p_1}$, where $\boldsymbol{p_1}, ..., \boldsymbol{p_n} \in \mathbb{R}^3$ denote predicted atom coordinates if they do exist. If $-\frac{1}{2}JDJ$ is positive semi-definite, then $M$ is also a symmetric positive semi-definite matrix and can be eigendecomposed as:
\begin{equation}
    M = U \Lambda U = \left(U \sqrt{\Lambda}\right) \left(\sqrt{\Lambda} U\right)^T = WW^T,
\end{equation} 
where $\Lambda=diag(\lambda_1,...\lambda_n)$ and $\lambda_1 \ge \lambda_2 \ge...\ge \lambda_n \ge 0$. Assuming that the first atom is placed at the origin of Cartesian space, $W$ exactly represents atom coordinates. Then, rotation and reflection can be naturally represented by applying an orthogonal matrix $Q \in \mathbb{R}^{n \times n}$. 

In short, given a molecule with its predicted distance matrix $D$, we compute the corresponding Gram matrix $M$ and then do eigendecomposition of $M$. If it yields all eigenvalues non-negative, then $D$ is a valid EDM. Otherwise, $D$ is invalid and cannot reconstruct atom coordinates in any Cartesian space. Based on this, the third metric \textsc{Validity} is proposed as:
\begin{equation}
    \textsc{Validity} = \frac{1}{|\mathbb{D}|} |\{D| D \textrm{\ is\ a\ valid\ EDM}, D \in \mathbb{D} \}|,
\end{equation}
where $\mathbb{D}$ denotes the set of all predicted distance matrices. Intuitively, \textsc{Validity} measures the percentage of molecules that are predicted with valid EDM. 

Even if the predicted distance matrix is a valid EDM, it may reconstruct atom coordinates in space higher than 3D due to the high rank of the Gram matrix $M$. To reconstruct valid 3D coordinates, the predicted $D$ has to be an EDM with a corresponding $M$ of rank three or less. Specifically, all eigenvalues of $M$ are expected to be non-negative, and three or fewer of them are positive. The last metric \textsc{Validity3D} is hence proposed to measure the percentage of molecules that are predicted with valid EDM and have corresponding atom coordinates in 3D space:
\begin{equation}
    \textsc{Validity3D} = \frac{1}{|\mathbb{D}|} |\{D| \textrm{Rank}(M) \le 3, M=\textrm{Gram}(D), D \in \mathbb{D}\}|.
\end{equation}


\section{The Proposed Benchmark}
The key innovations of our work are proposing to construct ground-state 3D molecular geometries via prediction methods, and presenting the Molecule3D dataset that empowers researchers to develop prediction methods. Based on the curated dataset, we provide a few benchmarking methods for ground-state 3D molecular geometry prediction in Section~\ref{sec: geo_pred}. Then in Section~\ref{sec:prop_pred}, we show an application example, namely molecular property prediction, based on the experimental results achieved in Section~\ref{sec: geo_pred}.


\subsection{3D Geometry Prediction}
\label{sec: geo_pred}
In this section, we benchmark a few methods for the novel task of ground-state 3D molecular geometry prediction. Specifically, we propose two deep learning methods, both based on the DeeperGCN-DAGNN model. Despite the same model architecture, these two methods are differently designed regarding the prediction target. We believe these benchmark methods can inspire more methods for the ground-state 3D molecular geometry prediction task.

\subsubsection{Input Features and Model Architecture}
To predict 3D molecular geometries, we use the DeeperGCN-DAGNN model \cite{liu2021fast}, which has been shown to be effective for learning molecular representations. To be specific, the overall network follows the DAGNN architecture \cite{liu2020towards}, which consists of $L$ transformation layers, $K$ propagation layers, and an adaptive adjustment mechanism, where $L$ and $K$ are hyper-parameters. We preserve the original design of DAGNN for the propagation layers and the adaptive adjustment mechanism since they can improve the quality of the learned representation by adaptively integrating the information of different receptive fields for each node. However, for the transformation layers, we instead adopt more expressive DeeperGCN \cite{li2020deepergcn} layers. The key component of a DeeperGCN layer is the softmax aggregator function. Besides, skip connections \cite{he2016deep} and pre-activation \cite{he2016identity} mechanisms are adopted in the DeeperGCN layer. For the sake of capturing long-range information, we also utilize the virtual node technique, which is shown to be helpful in molecular representation learning \cite{gilmer2017neural,hu2021ogb}.

We follow the way of the molecule datasets from the OGB \cite{hu2020open} benchmark to embed each node and edge in molecular graphs. Specifically, we use a 9-dimensional node feature vector describing the chemical properties of an atom such as its atomic number, chirality, and hybridization. A 3-dimensional edge feature vector is used for a bond to indicate its bond type, bond stereochemistry, and whether it is conjugated.

\subsubsection{Baseline Methods}
We provide two  methods based on the model architecture described in Section 4.1.1. The first baseline method considers the pairwise distances between atoms as the prediction targets. Formally, assuming that the model outputs $s$-dimensional vectors as node representations, the distance $d_{ij}$ between the $i$-th node and the $j$-th node is calculated by an element-wise max operation on these two node representations followed by a linear transformation. This process can be described by the following equations:
\begin{equation}
    \tilde{h}_{ij}=\circledast\left(h_i, h_j\right),
\end{equation}
\begin{equation}
    d_{ij} = W^T\tilde{h}_{ij}+b,
\end{equation}
where $h_i\in\mathbb{R}^s$ and $h_j\in\mathbb{R}^s$ denote node representations of the $i$-th and $j$-th node, respectively, $\circledast$ represents the element-wise max operation, and $W\in\mathbb{R}^s$ and $b\in\mathbb{R}$ are both trainable parameters.

In the second baseline method, we  directly predict the 3D coordinates of atoms. Specifically, the model outputs 3-dimensional vectors as node representations, and we consider these 3D representation vectors as the predicted 3D coordinates of atoms. Afterwards, we do not compute the difference between the predicted coordinates and the ground-truth coordinates as the loss function to train the model. Instead, we first compute the Euclidean distance $d_{ij}$ from the predicted 3D coordinates, then we compute the squared differences with respect to the  the corresponding ground-truth distances as the loss.

\subsubsection{Results and Discussion}

\begin{table}[t]
    \small
    \caption{Performance on 3D geometry prediction using both provided split types. For each metric, the best results are highlighted in bold numbers.}
    \center
    \subtable[Random split]{
        \addtolength{\tabcolsep}{-5pt}
        \begin{tabular}{lcccc|cccc}
            \toprule
            & \multicolumn{4}{c|}{Validation} & \multicolumn{4}{c}{Test} \\
            \cmidrule{2-9}
            & MAE & RMSE & Validity & Validity3D & MAE & RMSE & Validity & Validity3D \\
            \midrule
            RDKit ETKDG & 0.599 & 0.963 & \textbf{100\%} & \textbf{100\%} & 0.600 & 0.965 & \textbf{100\%} & \textbf{100\%} \\
            DeeperGCN-DAGNN + Distance & \textbf{0.482} & \textbf{0.749} & 1.71\% & 0.02\% & \textbf{0.483} & \textbf{0.753} & 1.69\% & 0.03\% \\
            DeeperGCN-DAGNN + Coordinates & 0.509 & 0.849 & \textbf{100\%} & \textbf{100\%} & 0.571 & 0.961 & \textbf{100\%} & \textbf{100\%} \\
            \bottomrule
            \end{tabular}
        \label{tab:geo_rand}}
    \subtable[Scaffold split]{
        \addtolength{\tabcolsep}{-5pt}
        \begin{tabular}{lcccc|cccc}
            \toprule
            & \multicolumn{4}{c|}{Validation} & \multicolumn{4}{c}{Test} \\
            \cmidrule{2-9}
            & MAE & RMSE & Validity & Validity3D & MAE & RMSE & Validity & Validity3D \\
            \midrule
            RDKit ETKDG & 0.551 & 0.893 & \textbf{100\%} & \textbf{100\%} & \textbf{0.532} & \textbf{0.880} & \textbf{100\%} & \textbf{100\%} \\
            DeeperGCN-DAGNN + Distance & \textbf{0.506} & \textbf{0.741} & 0.20\% & 0.00\% & 0.660 & 1.009 & 0.45\% & 0.02\% \\
            DeeperGCN-DAGNN + Coordinates & 0.617 & 0.930 & \textbf{100\%} & \textbf{100\%} & 0.763 & 1.176 & \textbf{100\%} & \textbf{100\%} \\
            \bottomrule
            \end{tabular}
        \label{tab:geo_scaffold}}
    \label{tab:geo}
\end{table}

The validation and test results are reported in Table~\ref{tab:geo}. Table~\ref{tab:geo_rand} and ~\ref{tab:geo_scaffold} show the results of using random split and scaffold split, respectively. For each split type, we also compare with the ETKDG algorithm \cite{riniker2015better} implemented in the RDKit \cite{rdkit} package. ETKDG is a conformer generation method developed upon the classical distance geometry (DG) \cite{blaney1994distance, havel1998distance} approach, but it integrates experimental torsion-angle preferences and additional basic knowledge such as flat aromatic rings. This method is much more efficient than DFT, but at the cost of accuracy. Note that RDKit ETKDG is not always successful in providing 3D geometries. After optimizing all molecules in our dataset by ETKDG, we find that 1311 and 1315 molecules fail in the validation and test set of the random split, respectively, and 1870 and 4434 molecules fail in the provided validation and test set of the scaffold split, respectively. 

From Table~\ref{tab:geo}, we can observe that our best baseline method achieves better prediction performance than RDKit ETKDG in terms of MAE and RMSE under the random split, but the performance of our methods is not as good as RDKit ETKDG for the scaffold split. This is not surprising since scaffold split requires the model to predict molecules whose scaffolds are very different from what the model has seen during training, as introduced in Section~\ref{subsec: splits}. Hence the prediction task becomes more challenging. We do not run experiments with various random seeds because we believe the Molecule3D dataset is sufficiently large to obtain relatively stable result. Besides, it is hard to repeat all experiments with various random seeds for such a large dataset.

In addition, the proposed two baseline methods are shown to achieve very different prediction performance even using the same model. Predicting the pairwise distances achieves smaller MAE and RMSE than predicting 3D coordinates directly. However, predicting the pairwise distances yields extremely low \textsc{Validity} and \textsc{Validity3D}. On the contrary, direct prediction of 3D coordinates ensures 100\% \textsc{Validity} and 100\% \textsc{Validity3D}, but results in larger error regarding pairwise distances. These two baseline methods indicate a trade-off between the error of distances and the validity of 3D coordinates. Ultimately, we pursue a solution to predict 3D geometries with both accurate pairwise distances and valid 3D coordinates.

\begin{wraptable}[10]{r}{7cm}\vspace{-0.4cm}
    \small
    \centering
    \caption{Time needed to predict 3D geometries over the entire test set. We time until coordinates are obtained.}
    \addtolength{\tabcolsep}{-5pt}
    \begin{tabular}{lc}
    \toprule
    & Time \\
    \midrule
    DFT & $>$150 days \\
    RDKit ETKDG & $\approx$7 hrs \\
    DeeperGCN-DAGNN + distance & $\approx$45 mins \\
    DeeperGCN-DAGNN + coordinates & $\approx$25 mins \\
    \bottomrule
    \end{tabular}
    \label{tab:time}
\end{wraptable}

We present a comparison of running time between different methods in Table~\ref{tab:time}. The results show that our methods are much more efficient than DFT and RDKit ETKDG. In other words, our methods can achieve reasonable geometry prediction accuracy while speeding up the computations compared with physics-based methods. This demonstrates that predicting ground-state 3D molecular geometries with deep learning approaches is a promising solution to reduce computational costs.

\subsection{Property Prediction}
\label{sec:prop_pred}

In Section~\ref{sec: geo_pred}, we have employed the DeeperGCN-DAGNN model to predict the ground-state 3D molecular geometries by either predicting the atom coordinates or the pairwise distances between all atoms. To further demonstrate the applicability of the predicted 3D geometries, we utilize the predicted 3D geometries obtained from Section~\ref{sec: geo_pred} to train a 3D graph network to predict the HOMO-LUMO gap. It is worthwhile emphasizing that Section~\ref{sec:prop_pred} simply serves as a sample downstream task of Section~\ref{sec: geo_pred}.

\subsubsection{Model Architecture}

According to the conclusion from Section~\ref{sec:validity}, a predicted EDM does not always be able to transform to 3D spatial coordinates because it may either be an invalid EDM or be associated with spatial coordinates of a space whose dimension is larger than 3. Hence, we do not employ more expressive 3D graph networks requiring 3D spatial coordinates for the computations of bond angles and torsion angles. Instead,  we adopt the simple SchNet \cite{schutt2017schnet} model to do quantum property prediction, which only requires the pairwise distances between atoms as inputs.

SchNet is a 3D graph network that can model the interaction between atoms in the molecule. It performs multiple message passing operations on the cutoff graph of the input 3D geometries. Specifically, any two atoms whose distances are lower than a specified threshold $\delta$ are connected in the cutoff graph. To incorporate the 3D information, SchNet proposes to use $n$ radial basis functions as edge features. Concretely, the edge features for the edge connecting the $i$-th and $j$-th node in the cutoff graph is:
\begin{equation}
	e_{ij}^k=\exp\left(-\gamma ||d_{ij}-\mu_k||_2^2\right), k=0,1,...,n
\end{equation}
where $\mu_k=\frac{k}{n}\delta$ and $d_{ij}$ is the distance between the $i$-th and $j$-th node. Here, $\delta$ and $n$ are both hyper-parameters. In our experiments, we use these proposed Gaussian basis functions as edge features and the same 9-dimensional node features as those used in the geometry prediction task.

We conduct HOMO-LUMO gap prediction experiments using our predicted geometries and ground-truth geometries, respectively. We also compare with predicting from molecular graphs. To make fair comparisons, we keep the network architecture the same and replace cutoff graphs by molecular graphs. The edge features are also replaced by the 3-dimensional edge features used in the geometry prediction task.

We evaluate the prediction performance by the mean absolute error (MAE) between the predicted properties and the ground-truth properties. Given a dataset of $m$ molecules whose HOMO-LUMO gaps are $\{\hat{y}_i\}_{i=1}^m$, and the predicted HOMO-LUMO gaps are $\{y_i\}_{i=1}^m$, where $y_i, \hat{y}_i\in\mathbb{R}$, the MAE is defined as:
\begin{equation}
	\mbox{MAE}\left(\{\hat{y}_i\}_{i=1}^m, \{y_i\}_{i=1}^m\right)=\frac{1}{m}\sum_{i=1}^{m}|\hat{y}_i-y_i|.
\end{equation}
A lower MAE indicates a better prediction performance.

\subsubsection{Results and Discussion}

\begin{table}[t]
    \small
    \caption{MAE performance on predicting the HOMO-LUMO gap using different inputs. Experiments are conducted with both provided split types. The best results are highlighted in bold numbers.}
    \center
    \subtable[Random split]{
        \addtolength{\tabcolsep}{-3pt}
        \begin{tabular}{lcc}
            \toprule
            Input & Validation & Test \\
            \midrule
            SMILES sequence & 0.1608 & 0.1624 \\
            Molecular graph & 0.2066 & 0.2062  \\
            Predicted 3D (Distance) & 0.1796 & 0.1813 \\
            Predicted 3D (Coordinates) & 0.1933 & 0.1941 \\
            RDKit ETKDG & 0.1728 & 0.1745 \\ 
            Ground-truth 3D & \textbf{0.0833} & \textbf{0.0833} \\
            \bottomrule
            \end{tabular}
        \label{tab:prop_rand}}
    \subtable[Scaffold split]{
        \addtolength{\tabcolsep}{-3pt}
        \begin{tabular}{lcc}
            \toprule
            Input & Validation & Test \\
            \midrule
            SMILES sequence & 0.1739 & 0.1430 \\
            Molecular graph & 0.2212 & 0.1890  \\
            Predicted 3D (Distance) & 0.2453 & 0.2000 \\
            Predicted 3D (Coordinates) & 0.2812 & 0.2371 \\
            RDKit ETKDG & 0.1730 & 0.1524 \\
            Ground-truth 3D  & \textbf{0.0883} & \textbf{0.0740} \\
            \bottomrule
            \end{tabular}
        \label{tab:prop_scaffold}}
    \label{tab:prop}
\end{table}



The prediction results in terms of MAE are presented in Table~\ref{tab:prop}. Using ground-truth 3D geometries are confirmed to achieve the best result. For random split, using predicted 3D geometries by all three baseline methods improves the prediction performance compared to using molecular graphs only. This result confirms our hypothesis that the predicted 3D geometries could benefit downstream property prediction tasks. Not surprisingly, it is harder to perform well on scaffold split, especially when both 3D geometry prediction and property prediction tasks employ scaffold split, which doubles the difficulty for our DeeperGCN-DAGNN based methods. 

\section{Conclusion and future works}

In this work, we propose Molecule3D, the very first benchmark for the ground-state 3D molecular geometry prediction problem. The main contributions of our work include: (1) curate a large dataset with more than 3.9 million molecules' ground-state 3D geometries and 4 quantum properties, (2) propose to construct 3D geometries via predictive methods, which is in a distinct direction from generative methods, (3) propose 4 metrics to evaluate the geometry prediction accuracy, and (4) provide AI-ready software tools for processing data and performing geometry prediction. In addition, we design two baseline methods based on the DeeperGCN-DAGNN \cite{liu2021fast} model to predict 3D geometries by either predicting pairwise distances between atoms or 3D coordinates of atoms. We show that compared with RDKit ETKDG \cite{riniker2015better} method, our methods can achieve comparable prediction accuracy but $7\times$ speedup. This demonstrates that deep learning methods are promising for significantly reducing computational costs of obtaining ground-state 3D molecular geometries. 

The example downstream prediction task of the HOMO-LUMO gap with a 3D GNN shows that the predicted 3D geometries can help the downstream task. However, the large gap of downstream performance between the ground-truth and predicted 3D geometries indicates ample room for improvement. Based on the proposed large dataset Molecule3D, we believe machine learning based prediction approaches have a great potential to predict the ground-truth 3D geometries accurately. Fast and accurately predicted 3D geometries will definitely advance studies of various real-world applications that rely on molecular structures. Therefore, we call on the machine learning community to work together on this challenging but meaningful task.


In the future, we will continue to present more metrics that may reveal the correlation between the predicted 3D geometry and the performance of downstream applications. Include but not limited to the root-mean-square deviation (RMSD), measures of bond angles and dihedral angles. In addition, we are currently processing an even larger dataset in which the ground-truth 3D geometries are optimized by the PM6 method \cite{nakata2020pubchemqc}. Compared to DFT, PM6 optimization is much faster but yields slightly inaccurate structures. With this dataset, we can do pre-training to enforce the model to learn more about 3D geometries. We plan to release this dataset soon. Furthermore, we will continue to develop prediction methods with novel designs to push up the performance of the ground-state 3D geometry prediction task. 

\begin{ack}
This work was supported in part by National Science Foundation grants IIS-1908198 and IIS-1908220.

\end{ack}

\bibliography{ref}
\bibliographystyle{plain}

\newpage
\appendix

\section{Appendix}

\subsection{Resource Availability and Licensing}

The proposed Molecule3D dataset includes raw molecule file, property file, processed PyG data file, and split files containing train/validation/test indices as described in Section~\ref{subsec: splits}. All mentioned files are available to download from \url{https://drive.google.com/drive/folders/1y-EyoDYMvWZwClc2uvXrM4\_hQBtM85BI?usp=sharing}. Since we provide the Molecule3D as a module of our MoleculeX software library, the MoleculeX package is highly recommended to be installed. Installation instructions can be found at \url{https://github.com/divelab/MoleculeX}. Details of Molecule3D, including data downloading, processing, splitting, training, and evaluation, can also be found at the Github homepage of MoleculeX. 

Molecule3D is licensed under a Creative Commons Attribution 4.0 International License. The authors bear all responsibility in case of violation of rights.

\subsection{Model Configurations}
\paragraph{Experimental Details.} We train our models for both 3D geometries prediction and property prediction on a single 11GB GeForce RTX 2080 Ti GPU. Our experiments are implemented with PyTorch 1.7.0 and PyTorch Geometric 1.7.0, and we use RDKit 2021.03.4. 

\paragraph{Model for 3D Geometry Prediction.} For our DeeperGCN-DAGNN models, we use $L=3$ DeeperGCN Layers and have $K=5$ in DAGNN. The number of hidden dimension is set to be 256. For the baseline method predicting pairwise distances, the output dimension is also 256. For the baseline method predicting 3D coordinates, the output dimension is set to be 3. The dropout rate is set to be 0. We train the model with the Adam optimizer \cite{kingma2015adam} for 60 epochs. The initial learning rate is 0.0001 and decays to 80\% at the 10th epoch. The batch size is 20. 

\paragraph{Model for Property Prediction.} For SchNet, we use 6 interaction blocks, and set number of hidden dimension and number of filters to be 256. The cutoff is set to be 10\AA. And we use mean pooling to obtain graph-level representations of molecules. We train the model with the Adam optimizer for 100 epochs. The initial learning rate is 0.0001 and decays to 96\% every 100,000 steps. The batch size is 32. 

\subsection{Data processing details} \label{sec:data_details}
The data provided by PubChemQC is not easily accessed. One needs to either use their database to query molecules one by one or download all raw data of size 2 TB. The downloaded raw data consists of millions of folders, and each contains one molecule. In every molecule folder, there are files including inp files of both ground-state and excited-state geometries and corresponding log files recording the optimization process. Note that inp files store atom coordinates only without any information about bonds. Thus, we use a mol file that is converted from the inp file by Open Babel. The mol file contains not only atom coordinates but also bonds and bond types. To make mol files more accessible by users, we read all mol files from each folder and integrate them into four SDF files. Moreover, we use the cclib package \cite{o2008cclib} to parse log files and extract quantum properties from massive information. Extracted properties are stored in a CSV file where each column saves one property. We find some molecules invalid during the above procedure due to SMILES conversion error and RDKit warnings, with sanitize problem, or with damaged log files. Thus we remove them and ensure molecules in SDF files and CSV file are in the same order.

The raw data of our Molecule3D consists of only one folder, including four SDF files, one CSV file, and four splits files. Raw data of Molecule3D has a size of 10 GB. It is tiny compared to the original data of PubChemQC that needs 2 TB. Hence, the large dataset becomes much more accessible to users who have a concern about memory. In addition to data size reduction, we also improve the loading efficiency. By reading our raw data, users can obtain 3,899,647 molecules with ground-state geometries in seconds. Based on what we have done, other users can save at least several weeks to download, understand, parse, clean and extract molecules from the PubChemQC database. 

\subsection{Datasheet for Molecule3D}

\subsubsection{Motivation for Datasheet Creation}

\textit{A. Why was the datasheet created? (e.g., was there a specific task in mind? was there a specific gap that needed to be filled?)}

Predicting ground-state 3D molecular geometries is very useful for molecular property analysis. However, there is no benchmark dataset specifically designed for this task. Hence, we create this dataset for anyone who is interested in this challenging problem.

\textit{B. What (other) tasks could the dataset be used for?}

The dataset can be used for (1) ground state 3D molecular geometry prediction, and (2) quantum property prediction.

\textit{C. Who funded the creation dataset?}

This work was supported in part by National Science Foundation grants IIS-1908198 and IIS-1908220.

\textit{D. Any other comment?}

N/A.

\subsubsection{Datasheet Composition}
\textit{A. What are the instances? (that is, examples; e.g., documents, images, people, countries) Are there multiple types of instances? (e.g., movies, users, ratings; people, interactions between them; nodes, edges)}

Each instance is a graph with nodes representing atoms and undirected edges representing chemical bonds. 

\textit{B. How many instances are there in total (of each type, if appropriate)?}

There are 3,899,647 molecules with great diversity included in Molecule3D.

\textit{C. What data does each instance consist of? ``Raw" data (e.g., unprocessed text or images)? Features/attributes? Is there a label/target associated with instances? If the instances related to people, are subpopulations identified (e.g., by age, gender, etc.) and what is their distribution?}

Raw data includes molecule files that can be read directly by RDKit, a CSV file storing quantum properties and pkl files storing split indices. We also provide a processed PyG data file for all molecules. By default, we follow the OGB benchmark to embed atom features and edge features. Users are allowed to design different features as needed. 

\textit{D. Is there a label or target associated with each instance? If so, please provide a description.}

Yes. The ground-truth 3D coordinates of atoms are labels of molecular graphs. Quantum properties including HOMO, LUMO, HOMO-LUMO gap and total energy are labels in the property prediction task.

\textit{E. Is any information missing from individual instances? If so, please provide a description, explaining why this information is missing (e.g., because it was unavailable). This does not include intentionally removed information, but might include, e.g., redacted text.}

No.

\textit{F. Are relationships between individual instances made explicit (e.g., users’ movie ratings, social network links)? If so, please describe how these relationships are made explicit.}

No.

\textit{G. Does the dataset contain all possible instances or is it a sample (not necessarily random) of instances from a larger set? If the dataset is a sample, then what is the larger set? Is the sample representative of the larger set (e.g., geographic coverage)? If so, please describe how this representativeness was validated/verified. If it is not representative of the larger set, please describe why not (e.g., to cover a more diverse range of instances, because instances were withheld or unavailable).}

As the number of existing molecules in the nature is very large, this dataset is a sample. We focus on small molecules, and provide a sufficiently large and representative dataset with over 3.9 million molecules.

\textit{H. Are there recommended data splits (e.g., training, devel- opment/validation, testing)? If so, please provide a description of these splits, explaining the rationale behind them.}

Yes, we provide both random split and scaffold split. Details about provided splits are written in Section~\ref{subsec: splits}.

\textit{I. Are there any errors, sources of noise, or redundancies in the dataset? If so, please provide a description.}

No.

\textit{J. Is the dataset self-contained, or does it link to or otherwise rely on external resources (e.g., websites, tweets, other datasets)? If it links to or relies on external resources, a) are there guarantees that they will exist, and remain constant, over time; b) are there official archival versions of the complete dataset (i.e., including the external resources as they existed at the time the dataset was created); c) are there any restrictions (e.g., licenses, fees) associated with any of the external resources that might apply to a future user? Please provide descriptions of all external resources and any restrictions associated with them, as well as links or other access points, as appropriate.}

Our dataset is self-contained. It does not rely on any external resources other than the Google Drive space storing the dataset.

\textit{K. Any other comments?}

N/A

\subsubsection{Collection Process}

\textit{A. What mechanisms or procedures were used to collect the data (e.g., hardware apparatus or sensor, manual human curation, software program, software API)? How were these mechanisms or procedures validated?}

We use rclone to download 2TB original data from the Google drive link provided by PubChemQC database. Then. we use python scripts to process and construct our dataset. 

\textit{B. How was the data associated with each instance acquired? Was the data directly observable (e.g., raw text, movie ratings), reported by subjects (e.g., survey responses), or indirectly inferred/derived from other data (e.g., part-of- speech tags, model-based guesses for age or language)? If data was reported by subjects or indirectly inferred/derived from other data, was the data validated/verified? If so, please describe how.}

Each instance is a molecular graph. The data is not directly observable. Atoms and bonds in molecules are represented as nodes and edges in graphs. For each molecular instance, its 3D geometry is represented by a list 3D vectors corresponding to 3D coordinates of atoms.

\textit{C. If the dataset is a sample from a larger set, what was the sampling strategy (e.g., deterministic, probabilistic with specific sampling probabilities)?}

We obtain molecules from the PubChemQC database whose source is the PubChem database. Molecules stored in PubChem are real molecules synthesized by the science and engineering communities and chemical vendors. PubChemQC provides only molecules whose 3D geometries are successfully optimized. 

\textit{D. Who was involved in the data collection process (e.g., students, crowdworkers, contractors) and how were they compensated (e.g., how much were crowdworkers paid)?}

The authors of the paper were the only ones involved in the data collection process.

\textit{E. Over what timeframe was the data collected? Does this timeframe match the creation timeframe of the data associated with the instances (e.g., recent crawl of old news articles)? If not, please describe the timeframe in which the data associated with the instances was created.}

It took about 1 week to collect all of the data.

\subsubsection{Data Preprocessing}

\textit{A. Was any preprocessing/cleaning/labeling of the data done (e.g., discretization or bucketing, tokenization, part-of-speech tagging, SIFT feature extraction, removal of instances, processing of missing values)? If so, please provide a description. If not, you may skip the remainder of the questions in this section.}

We have removed molecules with flawed structure or damaged raw file. For molecules remained in Molecule3D, there is no missing value or unsolved error. We leave each molecular graph in its original state where nodes/edges represent atoms/bonds.

\textit{B. Was the ``raw" data saved in addition to the preprocessed/cleaned/labeled data (e.g., to support unanticipated future uses)? If so, please provide a link or other access point to the “raw” data.}

PubChemQC database as the source of our Molecule3D is public and available at its homepage(\url{http://pubchemqc.riken.jp/}).

\textit{C. Is the software used to preprocess/clean/label the instances available? If so, please provide a link or other access point.}

No.

\textit{D. Does this dataset collection/processing procedure achieve the motivation for creating the dataset stated in the first section of this datasheet? If not, what are the limitations?}

Yes.

\textit{E. Any other comments}

No.

\subsubsection{Dataset Distribution}

\textit{A. How will the dataset be distributed? (e.g., tarball on website, API, GitHub; does the data have a DOI and is it archived redundantly?)}

We store all data files at Google Drive, and they are available for downloading from \url{https://drive.google.com/drive/folders/1y-EyoDYMvWZwClc2uvXrM4\_hQBtM85BI?usp=sharing}.

\textit{B. When will the dataset be released/first distributed? What license (if any) is it distributed under?}

We release the dataset under a Creative Commons Attribution 4.0 International License.

\textit{C. Are there any copyrights on the data?}

No.

\textit{D. Are there any fees or access/export restrictions?}

No.

\textit{E. Any other comments}

No.

\subsubsection{Dataset Maintenance}

\textit{A. Who is supporting/hosting/maintaining the dataset?}

The DIVE lab from the Department of Computer Science \& Engineering, Texas A\&M University.

\textit{B. Will the dataset be updated? If so, how often and by whom?}

As needed.

\textit{C. How will updates be communicated? (e.g., mailing list, GitHub)}

Since we provide the Molecule3D as a module of our MoleculeX software library, we will announce all related news on the Github homepage of MoleculeX (\url{https://github.com/divelab/MoleculeX}).

\textit{D. If the dataset becomes obsolete how will this be communicated?}

Since we provide the Molecule3D as a module of our MoleculeX software library, we will announce all related news on the Github homepage of MoleculeX (\url{https://github.com/divelab/MoleculeX}). 

\textit{E. Is there a repository to link to any/all papers/systems that use this dataset?}

No.

\textit{F. If others want to extend/augment/build on this dataset, is there a mechanism for them to do so? If so, is there a process for tracking/assessing the quality of those contributions. What is the process for communicating/distributing these contributions to users?}

There is not a mechanism to extend this dataset currently.

\subsubsection{Legal and Ethical Considerations}

\textit{A. Were any ethical review processes conducted (e.g., by an institutional review board)? If so, please provide a description of these review processes, including the outcomes, as well as a link or other access point to any supporting documentation.}

No.

\textit{B. Does the dataset contain data that might be considered confidential (e.g., data that is protected by legal privilege or by doctorpatient confidentiality, data that includes the content of individuals non-public communications)? If so, please provide a description.}

No.

\textit{D. Does the dataset relate to people? If not, you may skip the remaining questions in this section.}

No.

\textit{E. Does the dataset identify any subpopulations (e.g., by age, gender)? If so, please describe how these subpopulations are identified and provide a description of their respective distributions within the dataset.}

N/A

\textit{F. Is it possible to identify individuals (i.e., one or more natural persons), either directly or indirectly (i.e., in combination with other data) from the dataset? If so, please describe how.}

N/A

\textit{G. Does the dataset contain data that might be considered sensitive in any way (e.g., data that reveals racial or ethnic origins, sexual orientations, religious beliefs, political opinions or union memberships, or locations; financial or health data; biometric or genetic data; forms of government identification, such as social security numbers; criminal history)? If so, please provide a description.
}

N/A

\textit{H. Did you collect the data from the individuals in question directly, or obtain it via third parties or other sources (e.g., websites)?}

N/A

\textit{I. Were the individuals in question notified about the data collection? If so, please describe (or show with screenshots or other information) how notice was provided, and provide a link or other access point to, or otherwise reproduce, the exact language of the notification itself.}

N/A

\textit{J. Did the individuals in question consent to the collection and use of their data? If so, please describe (or show with screenshots or other information) how consent was requested and provided, and provide a link or other access point to, or otherwise reproduce, the exact language to which the individuals consented.}

N/A

\textit{K. If consent was obtained, were the consenting individuals provided with a mechanism to revoke their consent in the future or for certain uses? If so, please provide a description, as well as a link or other access point to the mechanism (if appropriate).}

N/A

\textit{L. Has an analysis of the potential impact of the dataset and its use on data subjects (e.g., a data protection impact analysis)been conducted? If so, please provide a description of this analysis, including the outcomes, as well as a link or other access point to any supporting documentation.}

N/A

\textit{M. Any other comments?}

No.

\end{document}